\pdfoutput=1
\documentclass[11pt]{article}

\usepackage[]{acl}

\usepackage{times}
\usepackage{latexsym}

\usepackage[T1]{fontenc}
\usepackage[utf8]{inputenc}

\usepackage{microtype}

\usepackage{mathtools}
\usepackage{graphics}
\usepackage{graphicx}
\usepackage{amsmath}
\usepackage{amsfonts,amssymb}
\usepackage{amssymb}
\usepackage{bbm}
\usepackage[ruled, linesnumbered]{algorithm2e}
\usepackage{amsmath}
\usepackage{booktabs}
\usepackage{multirow}
\usepackage{color}
\usepackage{makecell}
\usepackage{caption}
\usepackage{subcaption}
\usepackage{math_cmd}
\newcommand\myeq{\stackrel{\mathclap{\normalfont\mbox{def}}}{=}}

\title{$k$NN-NER: Named Entity Recognition with Nearest Neighbor Search}

\author{Shuhe Wang$^{1,2}$, Xiaoya Li$^{1}$, Yuxian Meng$^{1}$, Tianwei Zhang$^{3}$\\
{\bf Rongbin Ouyang$^{2}$, Jiwei Li$^{1,4}$, Guoyin Wang$^{5}$ }\\
$^1$Shannon.AI, $^2$Peking University,$^3$Nanyang Technological University\\
$^4$Zhejiang University, 
$^5$Amazon Alexa AI\\
\{shuhe\_wang, xiaoya\_li, yuxian\_meng, jiwei\_li\}@shannonai.com \\
tianwei.zhang@ntu.edu.sg, ouyang@pku.edu.cn, guoyiwan@amazon.com}

\begin{document}
\maketitle
\begin{abstract}
Inspired 
by recent advances in retrieval augmented methods in NLP~\citep{khandelwal2019generalization,khandelwal2020nearest,meng2021gnn}
, in this paper, we introduce a $k$ nearest neighbor NER ($k$NN-NER) framework, which augments the distribution
of entity labels 
by assigning $k$ nearest neighbors
retrieved 
from the training set.
This strategy 
makes the model more capable of handling long-tail cases, along with better few-shot learning abilities. 
$k$NN-NER requires no additional operation during the training phase, and by interpolating $k$ nearest neighbors search into the vanilla NER model, 
$k$NN-NER consistently outperforms its vanilla counterparts:
we achieve a new state-of-the-art F1-score of 72.03 (+1.25) on the Chinese Weibo dataset and 
improved results on 
a  variety of widely used NER benchmarks. 
Additionally, we show that $k$NN-NER can achieve comparable results to the vanilla NER model with 40\% less amount of training data.
\footnote{Code available at \url{https://github.com/ShannonAI/KNN-NER.}}
\end{abstract}

\section{Introduction}
Named Entity Recognition (NER) is an important problem in NLP, which refers to identifying named entities (\eg, person names, organizations, or locations) from a 
given
chunk of text. The most widely employed strategy for the NER task is to train a sequential labeling model based on a labeled dataset, 
and this model
learns to assign named entities to each token~\citep{chiu2016named, ma2016end, devlin2018bert}. This process of \textit{training} can be viewed as \textit{memorization}, in which the model iterates over the whole training set to memorize 
and generalize 
the most confident named entity assigned to the given word. This strategy of \textit{memorization} has difficulty in handling long-tail cases, and requires a large training set 
as sentence semantics get diverse and complicated 
~\citep{hammerton2003named,collobert2011natural,lample2016neural,chiu2016named,devlin2018bert,liu2019gcdt,shao2021self}.
%There are two key shortcomings for \textit{memorizing} the training set: (1) poor performance on cases that have low occurrences in the training set; (2) strong dependence on a large training set to memorize all kinds of data. \gw{add some citations here from some low-resource NER paper}

Motivated by recent progress in retrieval augmented methods \cite{khandelwal2019generalization,khandelwal2020nearest}, which have been 
successfully 
employed to handle similar issues in language generation, \ie
Language Modeling (LM) and  Neural Machine Translation (NMT), 
 we propose the $k$NN-NER framework for the NER task.
 $k$NN-NER first
  retrieves $k$ nearest neighbors from the cached training set. 
  Then, it computes  the distribution over labels by interpolating 
the distribution over labels output from 
    a vanilla NER model, 
    and weights for labels from similar examples in the training set, retrieved using token-level $k$NN search. 
In this way, we are able to     
  resolve the 
  long-tail 
  issue mentioned above: 
  by accessing the cached training examples through $k$NN search during inference, 
   similar cases (and their labels)  will shed  light on the test examples, which makes 
   memorizing the entire dataset unnecessary.

We conducted extensive experiments to evaluate the effectiveness of the proposed $k$NN-NER framework. 
We show that $k$NN-NER consistently outperforms its vanilla counterpart, which is based only on the 
distribution output from a vanilla tagging model and does not rely on similar examples in the training set. 
By applying $k$NN on the vanilla NER model with BERT~\citep{devlin2018bert} as the backbone, we 
are able to 
achieve a new state-of-the-art result 72.03 (+1.25) F1-score on the Chinese Weibo NER dataset and  results comparable to SOTA performances 
on a variety of widely applied NER benchmarks, including 
CoNLL03, OntoNotes5.0, 
 Chinese MSRA, and Chinese OntoNotes4.0. Additionally, our experiments show that $k$NN-NER can achieve comparable results to the vanilla NER model with 40\% less amount of training data.

\begin{figure*}[htb]
    \includegraphics[scale=0.315]{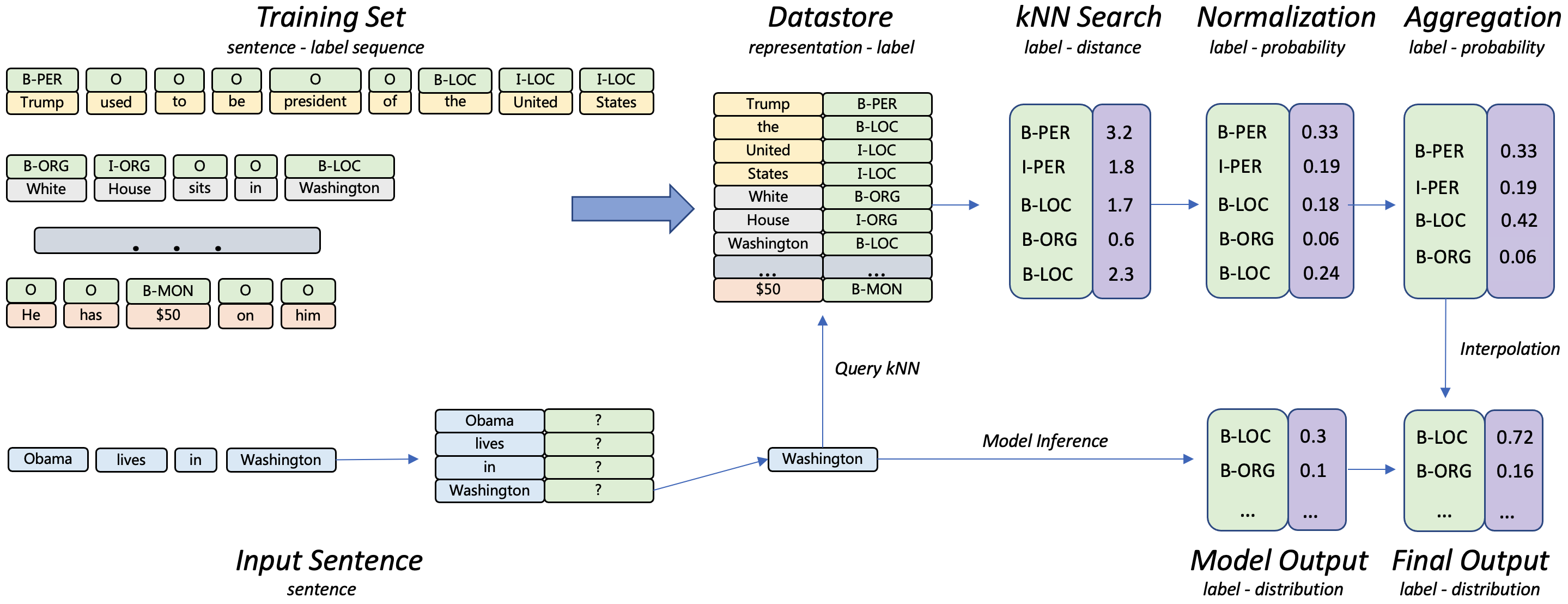}
    \caption{An example for the process of $k$NN-NER. The datastore contains a set of representation-label pairs, which are extracted from the hidden states of the vanilla NER model. By given an inference sentence: \textit{Obama lives in Washington}, suppose that at current test time $t$ we need to assign named entity to the word \textit{Washington}. The word representation of \textit{Washington} is used to query $k$ nearest neighbors from the datastore according to the similarity distance, and through the softmax function, the similarity distances are converted to $k$NN entity distribution. Interpolating the $k$NN distribution with the vanilla NER model distribution, we get the final distribution for the assigned named entities.}
    \label{fig:process}
\end{figure*}

\section{Related Work}

\paragraph{Retrieval Augmented Model} 
Retrieval augmented models additionally use the input to retrieve a set of relevant information to improve the model performance under the merit that \textit{an open-book exam is easier than a close-book exam}.
% “an open-book exam is easier than to a close-book exam”, retrieval augmented models additionally use the input to retrieve a set of relevant information.
Recent success on various NLP tasks has shown the effectiveness of retrieval augmented models in improving the quality of neural NLP models, such as language modeling~\citep{khandelwal2019generalization,meng2021gnn}, question answering~\citep{guu2020realm, lewis2020pre, lewis2020retrieval, xiong2020approximate},
text classification ~\cite{lin2021bertgcn}, 
dialog generation~\citep{fan2020augmenting, thulke2021efficient, weston2018retrieve} and neural machine translation~\citep{khandelwal2019generalization, meng2021fast,wang2021faster}.

\paragraph{Named Entity Recognition} Research on NER has a long history. \citet{hammerton2003named} first attempted to solve this problem using unidirectional LSTMs. \citet{collobert2011natural} presented a CNN-CRF structure and \citet{lample2016neural} combined the bidirectional LSTMs with CRFs. \citet{ma2016end} and \citet{chiu2016named} further added character feature via character CNN. Many works then focus on better improve the decoding structure: leveraging the context information \citep{liu2018empower, liu2019gcdt, lin2021asrnn, cui2019hierarchically}; interpolating latent variables \citep{lin2020enhanced, shao2021self}; transferring to CRF model \citep{ye2018hybrid, panchendrarajan2018bidirectional}; combining positive information \citep{dai2019joint}. Other researches like viewing the NER task as a machine reading comprehension(MRC) task also have made great performance \citep{li2019unified, li2019dice, gan2021dependency}.

\section{Proposed Method: $k$NN-NER}

\subsection{Background: Vanilla NER}
\paragraph{Sequence labeling for NER}
Given an input sentence $\xv=\{x_{1},...,x_{n}\}$ with length $n$, $\forall~ 1\leq i\leq n$, $x_{i}$ denotes the $i$-th word token within this sentence. We formalize the NER task as a sequence labeling task which assigns a label $y_{i}$ to each given word $x_{i}$. A training set with $N$ samples is then denoted by $\{\mathcal{X},\mathcal{Y}\}= \{(\xv^1,\yv^1), \cdots, (\xv^N, \yv^N)\}$, where $(\xv,\yv)$ is the text sequence and corresponding label sequence.
%A training sample is then formulated as $(\xv, \yv)=(\{x_{1},x_{2},...,{x_{n}}\}, \{y_{1},y_{2},...,{y_{n}}\})$.

For the vanilla NER model, we decompose the above sequence labeling task into two steps: (\emph{i}) using a text encoder to represent word tokens as high-dimensional vectors, and (\emph{ii}) classifying each high-dimensional vector into a named entity category. For step (\emph{i}), we use
masked language models, \eg, 
BERT\citep{devlin2018bert} and RoBERTa\citep{liu2019roberta},
as the  feature extractor. For a given word $x_i$, the output $\vh_i$ from the last layer of feature extractor is used as the contextualized word embedding vector, where $\vh_{i} \in \mathbb{R}^{m}$ with $m$ as the embedding dimension. 
Then for the step (\emph{ii}), we pass the word representation $\hv_i$ through a multi-layer perceptron (MLP), and obtain the distribution over the named entity vocabulary via a softmax operation:
\begin{equation}
\begin{aligned}
     p_{\text{NER}}(y_i|\vx, x_i) = \text{softmax}({\ \text{MLP}(\hv_{i}})).
\end{aligned}
\end{equation}

\subsection{$k$ Nearest Neighbor NER}
The key idea of the $k$NN-NER model is that it augments the process of classification during inference stage with a $k$ nearest neighbor retrieval mechanism. 
%Given a training dataset, we first create a datastore to cache the training dataset. Then the retrieval process directly extract relevant examples from the datastore during inference. 
As shown in Figure \ref{fig:process}, the $k$NN-NER process can be split into two parts: (\emph{i}) following the vanilla NER steps, \ie, extracting word representation $\vh$ and then assigning probability distribution $p_{\text{NER}}$ for each word in a given input sentence; and (\emph{ii}) finding the most similar contexts in the datastore and adjust the final entity distribution $p_{\text{final}}$ with a $k$NN-augmented entity distribution $p_{\text{KNN}}$.
In the following parts, we focus on two major components of $k$NN-NER framework: datastore construction and $k$NN entity probability interpolation.
%We now detail the part (2) in the following paragraphs: building datastore and interpolating $k$NN probability.

\paragraph{Building datastore} The datastore $\mathcal{D}$ consists of a set of \textit{key-value} pairs.
%and is used to cache examples to augment the assigned NER distribution $p_{\text{NER}}$.
Each \textit{key} is the contextualized word embedding of a word from a given sentence, and the corresponding \textit{value} is the name entity of that word in that sentence. 
%a high-dimensional vector representing the word of each sentence and the value is the corresponding named entity for that word. 
%Suppose that $S$ denotes a set of sentences and $L$ denotes the corresponding label set. Then the datastore can be formulated as:
Then the datastore $\mathcal{D}$ is formulated as:
\begin{equation}
\begin{aligned}
    \mathcal{D} \myeq{} \{\mathcal{K}, \mathcal{V}\} = &\{(\hv_i, y_{i}) | ~\forall x_{i} \in \xv, \forall y_{i} \in \yv,
    \\ &(\xv, \yv) \in \{\mathcal{X}, \mathcal{Y}\}\}.
\end{aligned}
\end{equation}
% \gw{this def needs further refine}
where $\hv_i$ is the contextualized representation of word $x_i$, $\mathcal{K}$ represents the set of \textit{keys} and $\mathcal{V}$ represents the corresponding \textit{value} set.
%where $s$ denotes a sentence in the full set $S$ and $l$ is the corresponding label sequence, $s_{i}$ denotes the word in sentence $s$ and $l_{i}$ is the named entity to $s_{i}$, $f(s, s_{i})$ represents a mapping function that changes the word $s_{i}$ in sentence $s$ to a high-dimensional vector. In this paper $f$ is the last layer result of BERT.\footnote{For the choice of mapping function $f$, experiments show that the results are better to use the last layer result of BERT rather than the layer result of MLP.}

\paragraph{$k$NN-augmented Entity Probability} Suppose that we have constructed the datastore $\mathcal{D}$. During inference time, for each word ${x_i}$ from a given input sentence $\xv$, our $k$NN-NER model first generates contextualized word embedding $\hv_{i}$ and distribution over the entire entity labels $p_{\text{NER}}(y_{i}|\xv, x_{i})$ for each word $x_{i}$.
%, where $1\leq i \leq n$, $n$ is the length of the input sentence and $l_{i}$ denotes the assigned named entity. 
Then for each word $x_{i}$, corresponding $\hv_{i}$ is used to query $k$ nearest neighbors set $\mathcal{N}$ from datastore $\mathcal{D}$ with $L^2$ Euclidean distance $d(\hv_i, \cdot)$ as similarity measure.
% squared $L^{2}$ distance $d(h_{i}, x)$, where $x$ denotes an item in datastore $D$. 

The retrieved named entity set is then converted into a distribution over the entire named entity vocabulary based on an RBF kernel output \citep{vert2004primer} of the distance to the original word embedding $\vh_i$. The probability of predicting the label as an entity $e_j$ is proportional to the summarization of kernel outputs from all \textit{values} in $\mathcal{N}$ which equal to $e_j$. 
% and the probability for each label is aggregated according to its occurrences while labels that do not appear in the retrieved set have zero probability:
% The retrieved named entity set is then converted into a distribution over the whole named entity vocabulary based on softmax, and the probability for each label is aggregated according to its occurrences while labels that do not appear in the retrieved set have zero probability:

\begin{equation}
\begin{aligned}
    p_{\text{kNN}}(y_{i}=&e_j|\xv, x_{i}) \varpropto\\ & \sum_{(\vk,v)\in \mathcal{N}} \mathbbm{1}_{v=e_j} \exp(\frac{-d(\vh_{i}, \vk)}{T})
\end{aligned}
\end{equation}
% \begin{equation}
% \begin{aligned}
%     p_{\text{kNN}}(l_{i}|s, s_{i}) \varpropto \sum_{(k_{j},v_{j})\in \mathcal{N}} \mathbbm{1}_{l_{i}=v_{i}} \exp(\frac{-d(h_{i}, k_{i})}{T})
% \end{aligned}
% \end{equation}
where $e_j$ represents the $j$th entity within the entity vocabulary and $T$ is a temperature parameter to flatten the distribution. Note that, for the labels that do not appear in the retrieved set, we always assign zero probability to these entities. 
Finally, we augment the pure NER distribution $p_{\text{NER}}(y_{i}|\xv, x_{i})$ with $p_{\text{kNN}}(y_{i}|\xv, x_{i})$ as:
% Finally, we augment the pure NER distribution $p_{\text{NER}}(l_{i}|s, s_{i})$ with $p_{\text{kNN}}(l_{i}|s, s_{i})$ as:
\begin{equation}
\begin{aligned}
    p_{\text{final}}(y_{i}|\xv,x_{i})=&\lambda p_{\text{NER}}(y_{i}|\xv, x_{i})\ +\\ &(1-\lambda) p_{\text{kNN}}(y_{i}|\xv, x_{i})
\end{aligned}
\end{equation}
% \begin{equation}
% \begin{aligned}
%     p(l_{i}|s,s_{i})=&\lambda p_{\text{NER}}(l_{i}|s, s_{i})\ +\\ &(1-\lambda) p_{\text{kNN}}(l_{i}|s, s_{i})
% \end{aligned}
% \end{equation}
where $\lambda$ makes a balance between $k$NN distribution and pure NER distribution.

\section{Experiments}
\subsection{Datasets}
% \begin{table}[th!]
%     \centering
%     \resizebox{.48\textwidth}{!}{
%     \begin{tabular}{llll}\toprule
%         \multicolumn{4}{c}{{\bf English CoNLL 2003}} \\\midrule
%         \textbf{Model} & \textbf{Precision} & \textbf{Recall} & \textbf{F1} \\\midrule
%         \multicolumn{4}{c}{{\it Base Model}} \\
%         BERT-Base~\citep{devlin2018bert} & 90.69 & 91.96 & 91.32 \\
%         BERT-Base+$k$NN & 91.50 & 91.58 & \textbf{91.54 (+0.22)} \\\midrule
%         \multicolumn{4}{c}{{\it Large Model}} \\
%         BERT-Large~\citep{devlin2018bert} & 91.54 & 92.79 & 92.16 \\
%         BERT-Large+$k$NN & 92.26 & 92.43 & \textbf{92.40 (+0.24)} \\\bottomrule
%         \multicolumn{4}{c}{{\bf English OntoNotes 5.0}} \\\midrule
%         \textbf{Model} & \textbf{Precision} & \textbf{Recall} & \textbf{F1} \\\midrule
%         \multicolumn{4}{c}{{\it Base Model}} \\
%         BERT-Base~\citep{devlin2018bert} & 85.09 & 85.99 & 85.54 \\
%         BERT-Base+$k$NN & 85.27 & 86.13 & \textbf{85.70 (+0.16)} \\\midrule
%         \multicolumn{4}{c}{{\it Large Model}} \\
%         BERT-Large~\citep{devlin2018bert} & 85.84 & 87.61 & 86.72 \\
%         BERT-Large+$k$NN & 85.92 & 87.84 & \textbf{86.87 (+0.15)} \\\bottomrule
%     \end{tabular}
%     }
%      \caption{Results for two English datasets: CoNLL 2003 and OntoNotes 5.0.}
%     \label{tab:result_on_english}
% \end{table}

We conduct experiments on commonly used English datasets and Chinese datasets. For English datasets, we use the widely 
used 
CoNLL2003 and OntoNotes 5.0 benchmarks. For Chinese datasets, we use OntoNotes 4.0, MSRA and Weibo NER. We adopt the general evaluation metric: span-level precision, recall and F1 score.  Dataset details are described at Appendix \ref{sec:appendix_datasets}.

\subsection{Experiment Results}
\begin{table}[th!]
    \centering
    \resizebox{.48\textwidth}{!}{
    \begin{tabular}{llll}\toprule
        \multicolumn{4}{c}{{\bf English CoNLL 2003}} \\\midrule
        \textbf{Model} & \textbf{Precision} & \textbf{Recall} & \textbf{F1} \\\midrule
        \multicolumn{4}{c}{{\it Base Model}} \\
        BERT-Base~\citep{devlin2018bert} & 90.69 & 91.96 & 91.32 \\
        BERT-Base+$k$NN & 91.50 & 91.58 & \textbf{91.54 (+0.22)} \\\midrule
        \multicolumn{4}{c}{{\it Large Model}} \\
        BERT-Large~\citep{devlin2018bert} & 91.54 & 92.79 & 92.16 \\
        BERT-Large+$k$NN & 92.26 & 92.43 & \textbf{92.40 (+0.24)} \\
        RoBERTa-Large~\citep{liu2019roberta} & 92.77 & 92.81 & 92.76 \\
        RoBERTa-Large+$k$NN & 92.82 & 92.99 & \textbf{92.93 (+0.17)} \\\bottomrule
        \multicolumn{4}{c}{{\bf English OntoNotes 5.0}} \\\midrule
        \textbf{Model} & \textbf{Precision} & \textbf{Recall} & \textbf{F1} \\\midrule
        \multicolumn{4}{c}{{\it Base Model}} \\
        BERT-Base~\citep{devlin2018bert} & 85.09 & 85.99 & 85.54 \\
        BERT-Base+$k$NN & 85.27 & 86.13 & \textbf{85.70 (+0.16)} \\\midrule
        \multicolumn{4}{c}{{\it Large Model}} \\
        BERT-Large~\citep{devlin2018bert} & 85.84 & 87.61 & 86.72 \\
        BERT-Large+$k$NN & 85.92 & 87.84 & \textbf{86.87 (+0.15)} \\
        RoBERTa-Large~\citep{liu2019roberta} & 86.59 & 88.17 & 87.37 \\
        RoBERTa-Large+$k$NN & 86.73 & 88.29 & \textbf{87.51 (+0.14)} \\\bottomrule
    \end{tabular}
    }
     \caption{Results for two English datasets: CoNLL 2003 and OntoNotes 5.0.}
    \label{tab:result_on_english}
\end{table}
\begin{table}[h!]
    \centering
    \resizebox{.48\textwidth}{!}{
    \begin{tabular}{llll}\toprule
        \multicolumn{4}{c}{{\bf Chinese OntoNotes 4.0}} \\\midrule
        \textbf{Model} & \textbf{Precision} & \textbf{Recall} & \textbf{F1} \\\midrule
        \multicolumn{4}{c}{{\it Base Model}} \\
        BERT-Base~\citep{devlin2018bert} & 78.01 & 80.35 & 79.16 \\
        BERT-Base+$k$NN & 80.23 & 81.60 & \textbf{80.91 (+1.75)} \\
        RoBERTa-Base~\citep{liu2019roberta} & 80.43 & 80.30 & 80.37 \\
        RoBERTa-Base+$k$NN & 79.65 & 82.60 & \textbf{81.10 (+0.73)} \\
        ChineseBERT-Base~\citep{sun2021chinesebert} & 80.03 & 83.33 & 81.65 \\
        ChineseBERT-Base+$k$NN & 81.43 & 82.58 & \textbf{82.00 (+0.35)} \\\midrule
        \multicolumn{4}{c}{{\it Large Model}} \\
        RoBERTa-Large~\citep{liu2019roberta} & 80.72 & 82.07 & 81.39 \\
        RoBERTa-Large+$k$NN & 79.87 & 83.17 & \textbf{81.49 (+0.10)} \\
        ChineseBERT-Large~\citep{sun2021chinesebert} & 80.77 & 83.65 & 82.18 \\
        ChineseBERT-Large+$k$NN & 81.68 & 83.46 & \textbf{82.56 (+0.38)} \\\bottomrule
        \multicolumn{4}{c}{{\bf Chinese MSRA}} \\\midrule
        \textbf{Model} & \textbf{Precision} & \textbf{Recall} & \textbf{F1} \\\midrule
        \multicolumn{4}{c}{{\it Base Model}} \\
        BERT-Base~\citep{devlin2018bert} & 94.97 & 94.62 & 94.80 \\
        BERT-Base+$k$NN & 95.34 & 94.64 & \textbf{94.99 (+0.19)} \\
        RoBERTa-Base~\citep{liu2019roberta} & 95.27 & 94.66 & 94.97 \\
        RoBERTa-Base+$k$NN & 95.47 & 94.79 & \textbf{95.13 (+0.16)} \\
        ChineseBERT-Base~\citep{sun2021chinesebert} & 95.39 & 95.39 & 95.39 \\
        ChineseBERT-Base+$k$NN & 95.73 & 95.27 & \textbf{95.50 (+0.11)} \\\midrule
        \multicolumn{4}{c}{{\it Large Model}} \\
        RoBERTa-Large~\citep{liu2019roberta} & 95.87 & 94.89 & 95.38 \\
        RoBERTa-Large+$k$NN & 95.96 & 95.02 & \textbf{95.49 (+0.11)} \\
        ChineseBERT-Large~\citep{sun2021chinesebert} & 95.61 & 95.61 & 95.61 \\
        ChineseBERT-Large+$k$NN & 95.83 & 95.68 & \textbf{95.76 (+0.15)} \\\bottomrule
        \multicolumn{4}{c}{{\bf Chinese Weibo NER}} \\\midrule
        \textbf{Model} & \textbf{Precision} & \textbf{Recall} & \textbf{F1} \\\midrule
        \multicolumn{4}{c}{{\it Base Model}} \\
        BERT-Base~\citep{devlin2018bert} & 67.12 & 66.88 & 67.33 \\
        BERT-Base+$k$NN & 70.07 & 67.87 & \textbf{68.96 (+1.63)} \\
        RoBERTa-Base~\citep{liu2019roberta} & 68.49 & 67.81 & 68.15 \\
        RoBERTa-Base+$k$NN & 67.52 & 69.81 & \textbf{68.65 (+0.50)} \\
        ChineseBERT-Base~\citep{sun2021chinesebert} & 68.27 & 69.78 & 69.02 \\
        ChineseBERT-Base+$k$NN & 68.97 & 73.71 & \textbf{71.26 (+2.24)} \\\midrule
        \multicolumn{4}{c}{{\it Large Model}} \\
        RoBERTa-Large~\citep{liu2019roberta} & 66.74 & 70.02 & 68.35 \\
        RoBERTa-Large+$k$NN & 69.36 & 70.53 & \textbf{69.94 (+1.59)} \\
        ChineseBERT-Large~\citep{sun2021chinesebert} & 68.75 & 72.97 & 70.80 \\
        ChineseBERT-Large+$k$NN & 75.00 & 69.29 & \textbf{72.03 (+1.23)} \\\bottomrule
    \end{tabular}
    }
     \caption{Results for three Chinese datasets: OntoNotes 4.0, MSRA and Weibo NER.}
    \label{tab:result_on_chinese}
\end{table}
\paragraph{The vanilla models} For the vanilla NER model, we choose BERT~\citep{devlin2018bert} and RoBERTa~\citep{liu2019roberta} for both English datasets and Chinese datasets, and ChineseBERT~\citep{sun2021chinesebert} only for Chinese datasets. 
Both base and large version of the vanilla NER model are used in our experiments. The details of implementation can be found in the original work, BERT~\citep{devlin2018bert}, RoBERTa~\citep{liu2019roberta} and ChineseBERT~\citep{sun2021chinesebert}. 
% All used vanilla models have two versions: base version with less parameters and large version with more parameters, and we recommend reading the original paper~\citep{devlin2018bert} and~\citep{sun2021chinesebert} for more model details.

\paragraph{Results} Table \ref{tab:result_on_english} and Table \ref{tab:result_on_chinese} show the results on English datasets and Chinese datasets respectively. \footnote{For two English datasets, we only interpolated $k$NN into pure BERT\citep{devlin2018bert} model, so the results are different from the reported ones. More details can be found in https://github.com/google-research/bert/issues/223}
% \gw{not quite sure about this part, why not we use the CRF one as it is quite commonly used?}
% Since the results reported in \citet{devlin2018bert} were enhanced with CRF\footnote{There are details for the results reported in \citet{devlin2018bert}. https://github.com/google-research/bert/issues/223}, our results are lower than the official Google BERT, while our key point is not reproducing the official results but observing the augment ability of $k$NN. 
We observe a significant improvement by interpolating $k$NN model across all tasks.  Especially on Chinese OntoNotes 4.0 and Chinese Weibo NER dataset, we observe an improvement of 1.75 and 1.63 respectively on F1-score based on BERT model.

\begin{figure}[h!]
    \centering
    \includegraphics[scale=0.31]{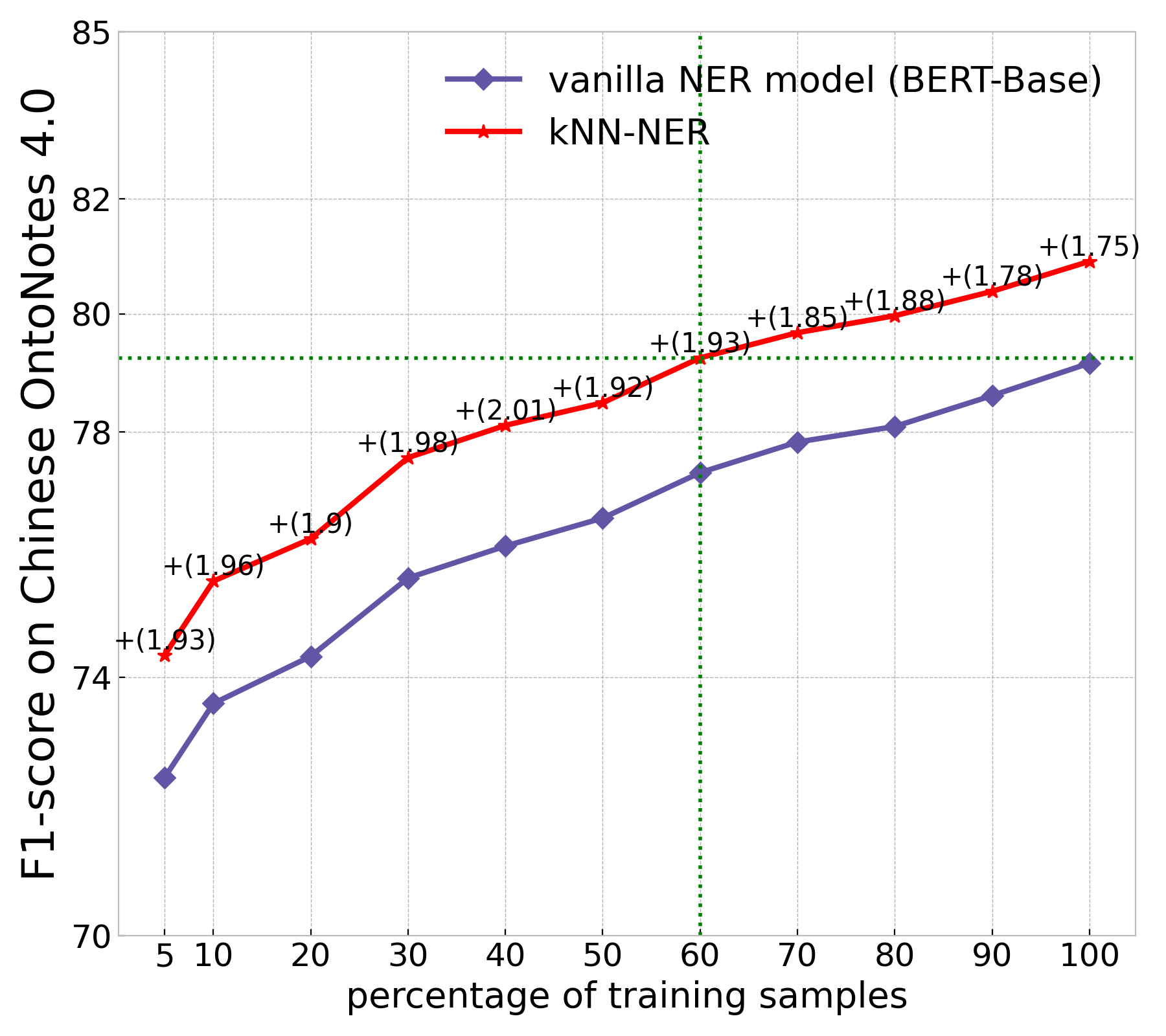}
    \caption{F1-score on Chinese OntoNotes 4.0 by varying the percentage of training set.}
    \label{fig:vary_percentage}
\end{figure}

\paragraph{Performance on low resource scenario} 
Empirically, we also observe that $k$NN-NER can achieve comparable results with much fewer training samples benefitting from direct access to the cached datastore. 
% Since $k$NN-NER can directly access the cached datastore, it may still achieve a comparable result by given a less training set. 
% We conducted experiments by varying the percentage of training set on Chinese OntoNotes 4.0 dataset to test this hypothesis.
On dataset Chinese OntoNotes 4.0, we conducted experiments by varying the percentage of training set while holding the full training set as the datastore for $k$NN search. 
Figure \ref{fig:vary_percentage} shows that without additional training and annotation, $k$NN-NER can still generate comparable result to the vanilla NER model with 40\% less amount of training data.
% \paragraph{Varying the size of training set} Since $k$NN-NER can directly access the cached datastore, it may still achieve a comparable result by given a less training set. We conducted experiments by varying the percentage of training set on Chinese OntoNotes 4.0 dataset to test this hypothesis. Figure \ref{fig:vary_percentage} shows the result, which $k$NN-NER can still generate comparable result to the vanilla NER model with 60\% amount of training data.

\paragraph{Effectiveness and Sensitivity of $k$}
To clearly observe the effectiveness of the hyperparameter $k$ during $k$NN search, we varied $k$ on dataset Chinese OntoNotes 4.0 with BERT as the vanilla NER model. 
% The results are shown in Figure \ref{fig:vary_k}. 
From Table \ref{tab:vary_k}, we observe that with the increase of $k$, the F1-score first increases and then keeps horizontal after $k$ reaches 256. 
% This is as expected, 
A larger $k$ can retrieve more informative neighbors from the cached datastore.
As $k$ continues increasing, the newly retrieved examples are less similar with the current input example and hence add ignorable change to the final performance. The steady performance with large enough $k$ values shows that our $k$NN-NER model is robust and not sensitive to choice $k$. 
% \gw{QQ: is this k much larger than the other knn framework? if so, we need to give an explanation} \shuhe{Answer: No, paper 1."NEAREST NEIGHBOR MACHINE TRANSLATION" 2."GENERALIZATION THROUGH MEMORIZATION: NEAREST NEIGHBOR LANGUAGE MODELS
% " 3."Fast Nearest Neighbor Machine Translation" all have the similar k.}

%\begin{figure}[h!]
%    \centering
%    \includegraphics[scale=0.31]{./ablation.png}
%    \caption{F1-score on Chinese OntoNotes 4.0 by varying the $k$NN parameter $k$. %\gw{same issue with the other plot, only one line is too few, consider add more lines or %4 small figs on more datasets}}
%    \label{fig:vary_k}
%\end{figure}

\begin{table}[th!]
    \centering
    \resizebox{.48\textwidth}{!}{
    \begin{tabular}{ll}\toprule
        \multicolumn{2}{c}{{\bf F1-score on Chinese OntoNotes 4.0}} \\\midrule
        \textbf{Varying $k$} & \textbf{F1-score} \\\midrule
        \text{The Vanilla NER Model} & 79.16 \\\midrule
         \text{+ by setting $k$=8} & 79.49(+0.33) \\
         \text{+ by setting $k$=16} & 79.67(+0.51) \\
         \text{+ by setting $k$=32} & 80.01(+0.85) \\
         \text{+ by setting $k$=64} & 80.53(+1.37) \\
         \text{+ by setting $k$=128} & 80.83(+1.67) \\
         \text{+ by setting $k$=256} & \textbf{80.91(+1.75)} \\
         \text{+ by setting $k$=512} & \textbf{80.91(+1.75)} \\\bottomrule
    \end{tabular}
    }
     \caption{F1-score on Chinese OntoNotes 4.0 by varying the $k$NN parameter $k$.}
    \label{tab:vary_k}
\end{table}

\section{Conclusion}
In this paper, we propose a new $k$NN-NER framework, which augments the generated distribution through assigning $k$ nearest neighbors from the cached training set. This strategy requires no additional operation on training phase. By applying $k$NN search on the vanilla NER model, we achieve a new state-of-the-art result 72.03 F1-score on Chinese Weibo NER dataset and comparable results on a variety of datasets, \eg, Chinese MSRA and Chinese OntoNotes 4.0. Additionally, our experiments show that $k$NN-NER can achieve comparable results to the vanilla NER model with only 60\% training data.

\bibliography{anthology,custom}
\bibliographystyle{acl_natbib}

\newpage
\appendix

\section{Experiments Datasets}
\label{sec:appendix_datasets}
\paragraph{English CoNLL2003.} CoNLL2003~\citep{sang2003introduction} is an English dataset containing four types of named entities: Location, Organization, Person and Miscellaneous, and we followed~\citep{li2019unified} leveraging protocols in~\citep{ma2016end} to process it.
\paragraph{English OntoNotes 5.0.} OntoNotes 5.0~\citep{pradhan2013towards} is an English dataset including 18 types of named entities: 11 types (e.g., Person, Organization) and 7 values (e.g., Date, Percent).
\paragraph{Chinese OntoNotes 4.0.} OntoNotes 4.0~\citep{pradhan2011proceedings} is a Chinese dataset with 18 types of named entities and all of them are extracted from news domain texts. Same as the CoNLL2003~\citep{sang2003introduction}, we followed~\citep{li2019unified} to process it.
\paragraph{Chinese MSRA.} MSRA~\citep{levow-2006-third} is a Chinese dataset collected from news domain texts. It contains three types of named entities and is used as shared task on SIGNAN backoff 2006.
\paragraph{Chinese Weibo NER.} Weibo NER~\citep{peng2015named} is a Chinese dataset drawn from the social media website Sina Weibo and includes four types of named entities.
%\section{Example Appendix}
%\label{sec:appendix}

%This is an appendix.

\end{document}